\documentclass[conference]{IEEEtran}
\IEEEoverridecommandlockouts
\pdfoutput=1
\usepackage{cite}
\usepackage{amsmath,amssymb,amsfonts}
\usepackage{algorithmic}
\usepackage{graphicx}
\usepackage{textcomp}
\usepackage{float}
\usepackage{xcolor}
\def\BibTeX{{\rm B\kern-.05em{\sc i\kern-.025em b}\kern-.08em
    T\kern-.1667em\lower.7ex\hbox{E}\kern-.125emX}}
\usepackage[skip=5pt plus1pt, indent=13pt]{parskip}
\begin{document}

\title{A Simplified Un-Supervised Learning Based Approach for Ink Mismatch Detection in Handwritten Hyper-Spectral
Document Images\\
}

\author{\IEEEauthorblockN{M. Farhan Humayun\IEEEauthorrefmark{1},
H. Waseem Malik\IEEEauthorrefmark{2} and A. Ahsan Alvi\IEEEauthorrefmark{3}}
\IEEEauthorblockA{Department of Electrical Engineering,
Institute of Space Technology\\
Islamabad, Pakistan\\
Email: \IEEEauthorrefmark{1}farhan.humayun2014@gmail.com,
\IEEEauthorrefmark{2}hassanwaseem009@gmail.com,
\IEEEauthorrefmark{3}ahmadahsanalvi@gmail.com}}
\maketitle

\maketitle

\begin{abstract}
Hyper-spectral imaging has become the latest trend in the field of optical imaging systems. Among various other applications, hyper-spectral imaging has been widely used for analysis of printed and handwritten documents. This paper proposes an efficient technique for estimating the number of different but visibly similar inks present in a Hyper-spectral Document Image.  Our approach is based on un-supervised learning and does not require any prior knowledge of the dataset. The algorithm was tested on the iVision HHID dataset and has achieved comparable results with the state of the algorithms present in the literature. This work can prove to be effective when employed during the early stages of forgery detection in Hyper-spectral Document Images.
\end{abstract}


\section{Introduction}
Hyper-spectral imaging (HSI) has gained a lot of popularity in recent times due to the capability of HSI sensors to record information in a number of bands corresponding to different wavelengths of the electromagnetic spectrum \cite{8314827}
. In contrast to the simple RGB senors, the HSI sensors can typically capture information in hundreds of bands. This information is helpful in determining the differences between features and objects that look similar to the naked human eye \cite{Land:71}. The applications of HSIs have been well studied and documented in detail in the literature \cite{article-min-zhu, article-yuen, Mateen2018, Li17, article-bold, Lu2014MedicalHI, article-edel}.

A recent field of research which has received considerable attention among the researchers is the Hyper-spectral Document Image (HSDI) analysis. This involves capturing and analysis of hyper-spectral images of printed and hand-written documents. These hard-copy documents are still extremely valuable despite many hurdles in their storage, management and retrieval. Specific examples include hard-copy evidences involved in court room proceedings as well as old manuscripts having historical significance. Among different applications of HSI, one particular application is that it can differentiate among various types of materials present in the image based on each material's spectral reflectance properties. The core idea is that each type of material has its own specific spectral response curve corresponding to various bands of the hyper-spectral image, therefore different materials can be differentiated by analysing the spectral response curves of these materials. This property of hyper-spectral imaging systems, enables them to be used as a non-destructive tool for document analysis including forgery detection \cite{8d8871b7b272427ea8b4d12e3e256ea4}, restoration of old and de-graded scripts \cite{10.1007/s11042-017-5564-2},  ink mismatch detection \cite{6628744, 8270134}, signature extraction \cite{article-kashif}, writer identification \cite{8945886}, readability enhancement \cite{38d62691d8dc4a1a83da13335fde6bac} and  estimating the age of old manuscripts \cite{Rahiche2020HistoricalDD}.

This paper describes an automated technique for detecting the number of different inks present in a hyper-spectral document image using an unsupervised learning algorithm. A lot of work has also been done on ink mis-match detection using supervised learning algorithms such as Deep Convolutional Neural Networks (DCNNs) \cite{8978076, article-khan, Devassy2019InkCU}. However, a major limitation of the supervised learning algorithms is that they need prior labeled dataset i.e. the number and type of inks present in the document for training purposes which makes them difficult to use in practical scenarios. Previous works involving unsupervised learning algorithms for HSDI analysis include K-means clustering, C-means clustering, Support Vector Machines (SVMs) and Principal Component Analysis (PCA) \cite{10.1007/11669487_4, article-qur, article-zohaib}. 

The rest of the paper is divided into following sections: Section II describes the proposed methodology in detail including the design decisions for selection of programming framework/libraries and the choice of unsupervised learning algorithm. Section III describes the experimental results. Section IV contains the conclusion along with limitations of the current work as well as future research directions.

\section{Methodology}

\begin{figure}[H]
    \centering
    \includegraphics[width=5cm]{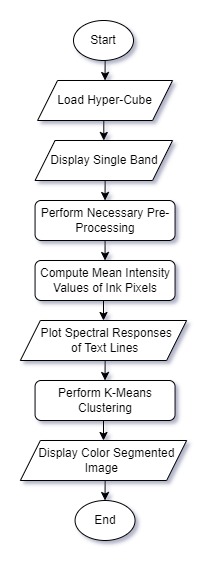}
    \caption{Flow chart of the Proposed Methodology}
    \label{fig:flow}
\end{figure}

We propose a simplified and automated approach for classifying different types of inks present in hand-written HSDIs. The overall procedure involves various steps including pre-processing of hyper-spectral cubes, generation of spectral signatures of the hand-written text and implementation of the un-supervised learning algorithm for displaying the color segmented image as the final output. This section discusses the technical details of implementation along with the description of the design decisions regarding the choice of programming framework and the unsupervised algorithm. A flow chart of the overall procedure for automated detection of number of inks in HSDIs is given in Fig \ref{fig:flow}.

\subsection{Design Decisions}

\subsubsection{Programming Framework}
For the easy of implementation and a strong built-in support for handling and manipulating hyper-spectral images, Python was chosen as the underlying programming language to implement the proposed algorithm. The complete implementation was carried out using the following four libraries of Python:
\begin{itemize}
\item Spectral (for handling Hyper-spectral cubes).
\item Numpy (for array manipulation).
\item OpenCV (for manipulating RGB and grey-scale images).
\item Matplotlib (for plotting of graphs and figures).
\end{itemize}

\subsubsection{Choice of Un-supervised Algorithm}
Among a wide variety of un-supervised algorithms, including SVM, PCA, K-means, C-means clustering etc., K-means was chosen because of its simplicity and efficiency in determining accurate clusters from un-labeled data. Another reason for selecting K-means algorithm was that the Spectral library contains a built-in function specifically tailored for implementation of k-means clustering algorithm on hyper-spectral datasets.

\subsection{Data Pre-processing}
A number of steps were involved in the pre-processing of the hyper-spectral cubes before they could be used for plotting the spectral signatures of text pixels as well as implementation of the un-supervised pattern recognition algorithm. These steps are discussed below:

\subsubsection{Cropping}
The first step was to crop the original hyper-spectral cube to retain only the relevant area where most of the text pixels lie. This was achieved using simple slicing operations of the numpy arrays containing the hyper-spectral cube. A single band was used to display a grey-scale image for verifying the results of the cropping procedure.
Similarly, separate cubes for each line of text were also cropped and saved in different arrays. These were used for plotting the spectral signatures of each line of text.

\subsubsection{Binary Thresholding}
Binary thresholding was applied on a single band to convert the grey-scale image displayed during cropping into a binary image containing only black ink pixels on white background. For estimating the value of threshold, histogram of the single band image was computed and the threshold value was picked based on visual analysis of the distribution of pixels in the image. The same threshold was also applied on cropped images pertaining to each line separately. 

\subsubsection{Boolean Masking}
Boolean mask was applied on the binary image to know the positions of foreground pixels i.e. ink pixels only. The Boolean mask returned the value of \emph{True} in place of ink pixels and \emph{False} for the background pixels. Similar masks were applied for each individual lines to know the exact positions ink pixels in each line.

\subsection{Plotting Spectral Signatures of Ink Pixels}
The boolean masks pertaining to each individual lines were applied to the cropped cubes of each line to extract ink pixels. Once the ink pixels were extracted, the means of pixels intensities of each line were calculated across various bands present in the hyper-spectral cube. These mean intensity values of each individual line were plotted against the number of bands / wavelengths to generate spectral response curves of every line of text.

\subsection{Implementing K-Means clustering}
For the purpose of implementing clustering algorithm, we could not simply extract the ink pixels of the whole document, because i9n that case their spatial information would be lost and it would not be possible to display the results of clustering algorithm in the form an image. To rectify this issue we also included the background pixels but suppressed the intensities values of of background pixels to either zero (black background) or one (white background). The main idea behind this approach is that due to the constant value of background pixels, the K-means clustering algorithm allots them a single cluster rather then making further classes within the background pixels. It just focuses on the pixels which have variations i.e. the foreground ink pixels and the spatial context of the ink pixels is also retained. After that, the built-in function of Spectral library for k-means algorithm is implemented on the cropped hyper-spectral cube with suppressed background. The output of K-means is displayed in the form of a plot representing different classes generated by the algorithm.  

\subsection{Generating Color Segmented Image as Final Output}
The output clusters from k-means algorithm is used to display the final output image in the form of an RGB image showcasing lines of text written with different inks with different colors. A built-in function from the Spectral library is used to visualize the resultant color segmented image.

\section{Experimental Results and Analysis}
This section discusses in detail the experimental setup including dataset used as well as the system specifications. The results of each step as discussed in the methodology section are also displayed in chronological order. 

\subsection{iVision HHID Dataset}

A single hyper-spectral cube from the Hand-written Hyper-spectral Images Dataset (HHDI) \cite{ISLAM2022107964} was used for implementing the proposed methodology. The dataset contains hyper-spectral cubes, each of resolution 650 x 512 pixels and 149 bands ranging between wavelengths of approximately 478nm to 901nm. The hyper-spectral cube used to carry out the experiment contained twelve lines of same text: "A quick brown fox jumps over the lazy dog" written with different kinds of blue inks but each individual line was written with a single ink. Fig 2 displays bands 1, 10 and 30 for visual inspection of the information contained in the hyper-spectral cube.

\begin{figure}[H]
    \centering
    \includegraphics[width=8.5cm]{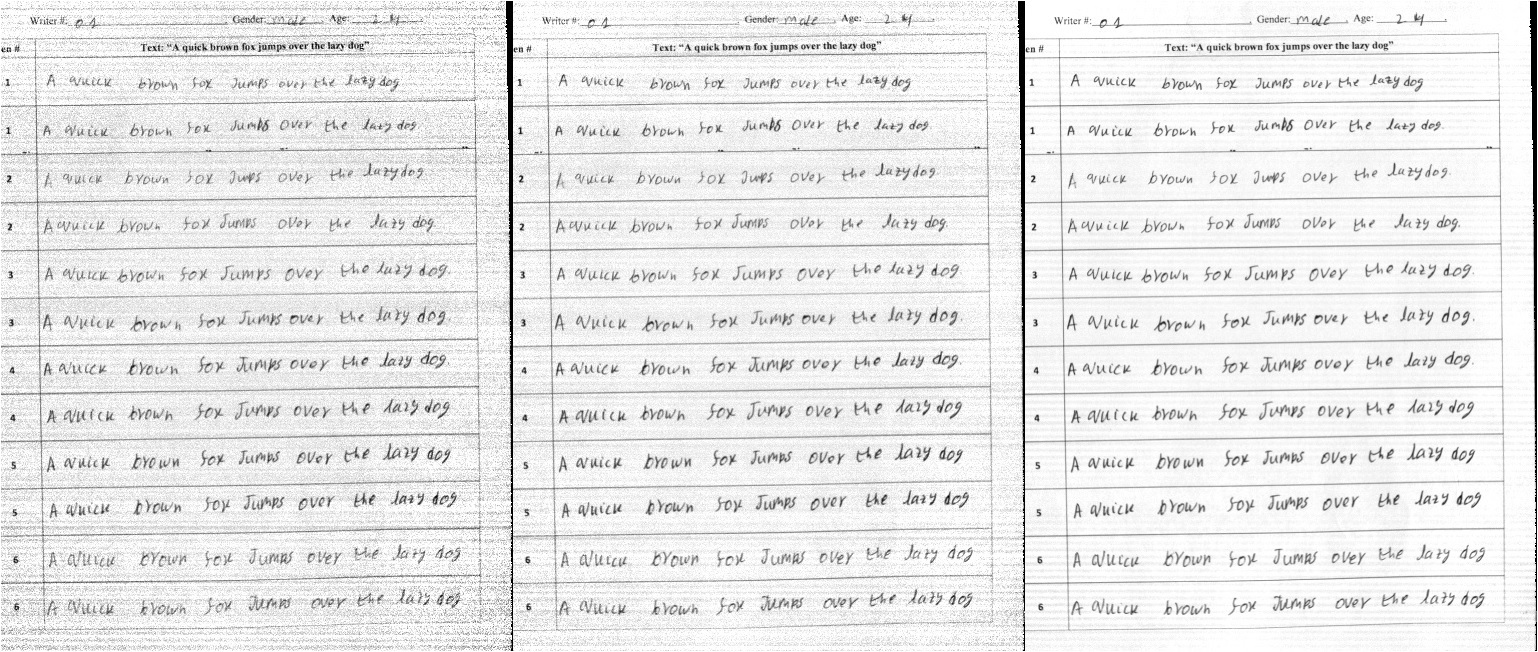}
    \caption{Band 1, 10 and 30 as Separate Grey-Scale Images}
\end{figure}

\subsection{Hardware System Specifications}
The proposed ink detection procedure was implemented on a hardware with following specifications:
\begin{itemize}
\item CPU: Intel Core i3 @ 2.30 GHz.
\item RAM: 6 GB DDR3.
\item Hard Drive: 512 GB SSD.
\item OS: Windows 10 Pro 64-bit.
\end{itemize}

\subsection{Intermediate Outputs in the Pre-Processing Stage}
The full scale cropped image is shown in Fig 3. The unwanted details from left, right, top and bottom of the original image were removed to keep only the relevant area containing hand-written text.

\begin{figure}[H]
    \centering
    \includegraphics[width=4cm]{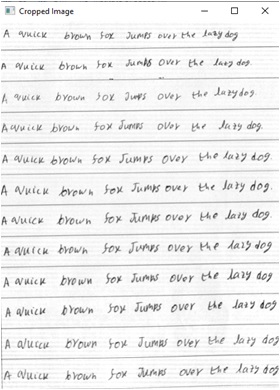}
    \caption{Result of performing cropping operations on the image}
\end{figure}
 
The histogram of band 30 is displayed in Fig 4. The values on the x-axis of the histogram are used to estimate the threshold value for generating binary image.

\begin{figure}[H]
    \centering
    \includegraphics[width=8cm]{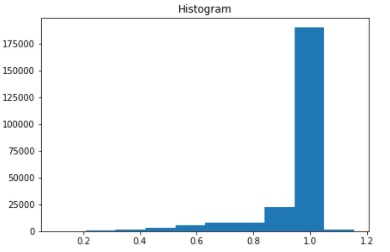}
    \caption{Histogram of grey-scale image corresponding to band no. 30 of the hyper-cube}
\end{figure}

After applying thresholding function, the corresponding black and white image is shown in Fig 5. Similarly results of separate binary images generated for each individual lines of text is displayed in Fig 6.

\begin{figure}[H]
    \centering
    \includegraphics[width=4cm]{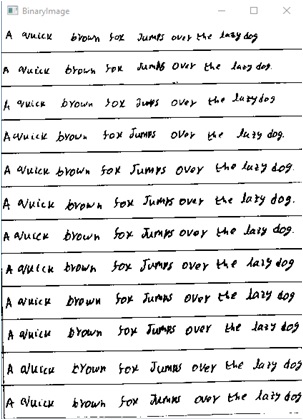}
    \caption{Result of applying binary thresholding on cropped image}
\end{figure}

\begin{figure}[H]
    \centering
    \includegraphics[width=8cm]{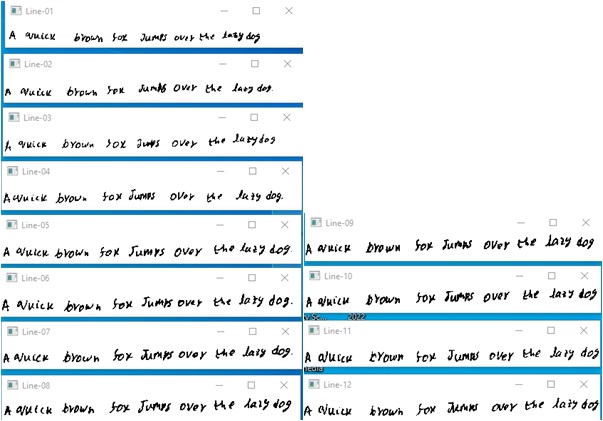}
    \caption{Result of cropping and binary thresholding on individual lines of text}
\end{figure}

Fig 7 shows the result of applying Boolean masking on the cropped image. Positions of all the ink pixel are returned as \emph{True} values, whereas background pixels are returned as \emph{False} values. Similar masks were generated for each individual lines of text as well.

\begin{figure}[H]
    \centering
    \includegraphics[width=4cm]{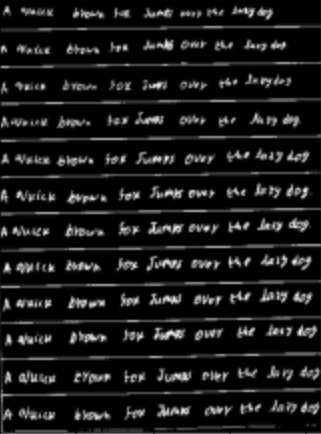}
    \caption{Result of applying Boolean mask on the binary image}
\end{figure}

\subsection{Spectral Responses of Text Pixels}
After applying the boolean masks of all lines to their corresponding hyper-cubes, the mean intensity values were computed for each individual line of text and plotted against all the 149 different wavelengths available in the cube. The resulting spectral response graphs are displayed in Fig 8.

\begin{figure}[H]
    \centering
    \includegraphics[width=8.5cm]{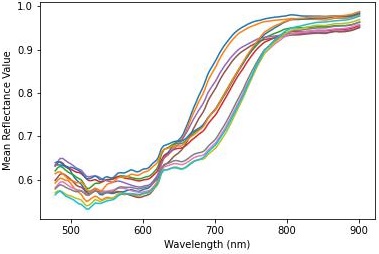}
    \caption{Mean Spectral Responses of 12 lines of text}
    
\end{figure}
Based on the visual analysis of spectral responses, graphs with similar trends were plotted separately. Figs 9 - 13 show the spectral graphs pertaining to different lines of text with similar trends. Based on these graphs, it can be inferred that there are probably five major clusters pertaining to five different inks present in the document.
\begin{figure}[H]
    \centering
    \includegraphics[width=7cm]{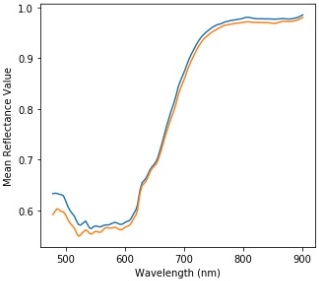}
    \caption{Spectral Responses pertaining to lines 1 and 2 of text}
    
\end{figure}

\begin{figure}[H]
    \centering
    \includegraphics[width=7cm]{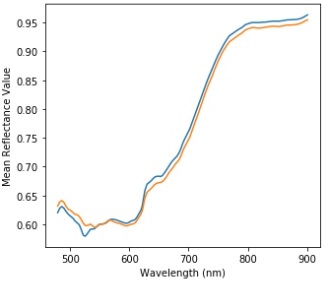}
    \caption{Spectral Responses pertaining to lines 3 and 4 of text}
    
\end{figure}

\begin{figure}[H]
    \centering
    \includegraphics[width=7cm]{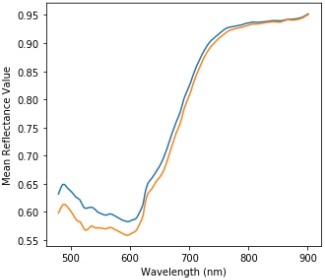}
    \caption{Spectral Responses pertaining to lines 5 and 6 of text}
    
\end{figure}

\begin{figure}[H]
    \centering
    \includegraphics[width=7cm]{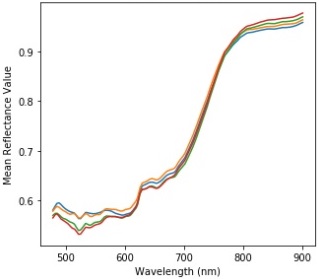}
    \caption{Spectral Responses pertaining to lines 7, 8, 9 and 10 of text}
    
\end{figure}

\begin{figure}[H]
    \centering
    \includegraphics[width=7cm]{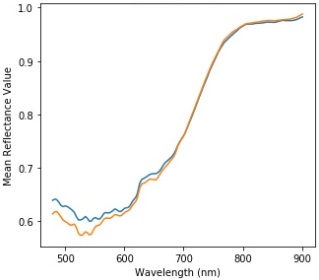}
    \caption{Spectral Responses pertaining to lines 11 and 12 of text}
    
\end{figure}

\subsection{Results from K-means Clustering}
For the purpose of automatic clustering, K-means clustering was implemented as the next step in the proposed approach. The boolean masked image after suppression of background pixels was fed to the K-means clustering algorithm. After experimenting with different values of K, the final value of K was chosen to be 7 which produced best results. For the hyper cube used in the experiments, the algorithm converged after 84 iterations. Fig. 14 shows the plots of spectral classes that have been identified by the K-means clustering. 

It is interesting to see that there are 7 plots in Fig. 15. The bottom-most straight plot with reflectance value of 0 corresponds to the suppressed background pixels, whereas the top-most plot corresponds to the flat horizontally printed lines in-between the text. During the pre-processing stage, we did not remove these intermittent lines to avoid complexity. The rest of the five plots correspond to the five major ink clusters present in the HSDI.

\begin{figure}[H]
    \centering
    \includegraphics[width=8cm]{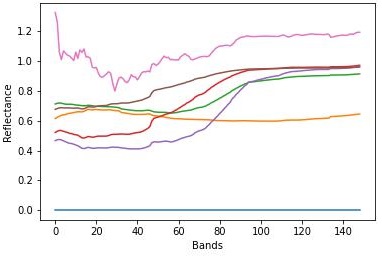}
    \caption{Spectral Classes from K-means Clustering}
    
\end{figure}

\subsection{The Final Color Segmented Images}
The color segmented RGB images were generated using the spectral classes from the output of the K-means clustering algorithm. Fig 14 shows the final resultant images. Left image is generated when the background pixels are assigned the value 1, whereas in the right image, the background pixels are allotted a constant value of 0. These output images confirm the presence of at-least five major clusters of inks in the documents. The output is also summarized in Table 1.

\begin{figure}[H]
    \centering
    \includegraphics[width=8cm]{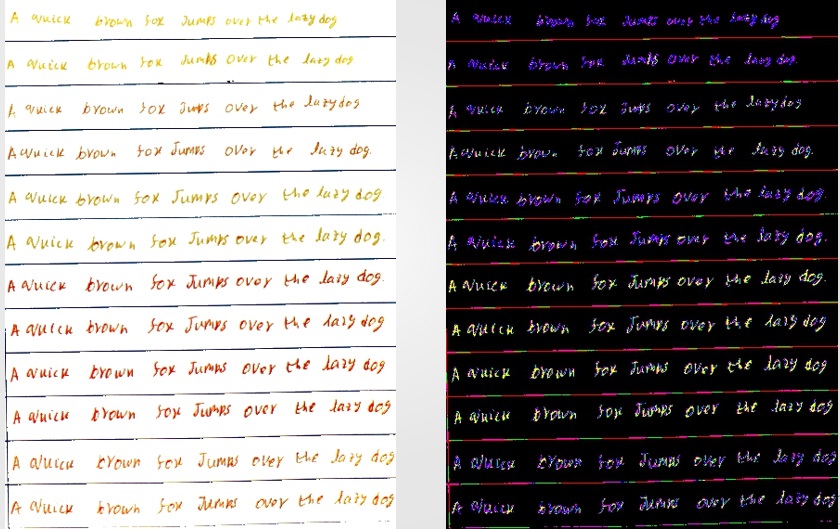}
    \caption{Color Segmented Images}
    
\end{figure}

\begin{table}[htbp]
\caption{Text Lines Corresponding to Different Clusters}
\renewcommand{\arraystretch}{1.5}

\begin{center}
\begin{tabular}{|c|c|c|}
\hline
\textbf{Sr. No.}& \textbf{Cluster No.}& \textbf{Text Line No.} \\
\hline
1& C-1& 1,2 \\
\hline
2& C-2& 3,4 \\
\hline
3& C-3& 5,6 \\
\hline
4& C-4& 7,8,9,10 \\
\hline
5& C-5& 11,12 \\
\hline
\end{tabular}
\label{tab1}
\end{center}
\end{table}

\section{Conclusion}
In this paper, we have proposed a simplified yet efficient approach for ink mismatch detection in HSDIs. After the necessary pre-processing, the spectral responses of the ink pixels are plotted which provide an estimate of the different number of inks present in the document. After that K-means clustering is implemented on the hyper-spectral cube to form various clusters of inks. The results from K-means algorithm confirm the initial estimate of five major clusters from visual analysis of the the spectral curves. This work can prove to be of significant value for quickly estimating the degree of ink mismatch during the early phases of forgery detection in HSDIs.  

The advantage of our approach is that it is robust and can be easily used for estimating the number of inks present in an HSDI. Our approach does not require prior knowledge of the dataset or pre-training with labeled examples. A limitation of the approach is that we did not remove the flat horizontally printed lines in between the text pixels. This caused an extra cluster to be produced in the output of K-means. Another limitation is that the actual ground truth contains six different inks in the HSDI, whereas, our algorithm was able to detect only five major clusters. In future work, we aim to improve the results as well as remove the extra cluster from the output of K-means clustering. We also aim to explore other un-supervised algorithms including C-means, PCA and SVM etc and relevant HSDI datasets for comparative analysis. 

\section{Acknowledgment}

We are thankful to Dr. Khurram Khurshid (Institute of Space Technology) for his guidance and expertise related to Hyper-Spectral Document Image (HSDI) analysis.

\bibliographystyle{IEEEtran}
\bibliography{Reference.bib}

\begin{thebibliography}{10}
\providecommand{\url}[1]{#1}
\csname url@samestyle\endcsname
\providecommand{\newblock}{\relax}
\providecommand{\bibinfo}[2]{#2}
\providecommand{\BIBentrySTDinterwordspacing}{\spaceskip=0pt\relax}
\providecommand{\BIBentryALTinterwordstretchfactor}{4}
\providecommand{\BIBentryALTinterwordspacing}{\spaceskip=\fontdimen2\font plus
\BIBentryALTinterwordstretchfactor\fontdimen3\font minus
  \fontdimen4\font\relax}
\providecommand{\BIBforeignlanguage}[2]{{%
\expandafter\ifx\csname l@#1\endcsname\relax
\typeout{** WARNING: IEEEtran.bst: No hyphenation pattern has been}%
\typeout{** loaded for the language `#1'. Using the pattern for}%
\typeout{** the default language instead.}%
\else
\language=\csname l@#1\endcsname
\fi
#2}}
\providecommand{\BIBdecl}{\relax}
\BIBdecl

\bibitem{8314827}
M.~J. Khan, H.~S. Khan, A.~Yousaf, K.~Khurshid, and A.~Abbas, ``Modern trends
  in hyperspectral image analysis: A review,'' \emph{IEEE Access}, vol.~6, pp.
  14\,118--14\,129, 2018.

\bibitem{Land:71}
\BIBentryALTinterwordspacing
E.~H. Land and J.~J. McCann, ``Lightness and retinex theory,'' \emph{J. Opt.
  Soc. Am.}, vol.~61, no.~1, pp. 1--11, Jan 1971. [Online]. Available:
  \url{http://opg.optica.org/abstract.cfm?URI=josa-61-1-1}
\BIBentrySTDinterwordspacing

\bibitem{article-min-zhu}
M.~Zhu, D.~Huang, X.~Hu, W.~Tong, B.~Han, J.~Tian, and H.~Luo, ``Application of
  hyperspectral technology in detection of agricultural products and food: A
  review,'' \emph{Food Science and Nutrition}, vol.~8, 09 2020.

\bibitem{article-yuen}
P.~Yuen and M.~Richardson, ``An introduction to hyperspectral imaging and its
  application for security, surveillance and target acquisition,''
  \emph{Imaging Science Journal, The}, vol.~58, pp. 241--253, 10 2010.

\bibitem{Mateen2018}
\BIBentryALTinterwordspacing
M.~Mateen, J.~Wen, Nasrullah, and M.~A. Akbar, ``The role of hyperspectral
  imaging: A literature review,'' \emph{International Journal of Advanced
  Computer Science and Applications}, vol.~9, no.~8, 2018. [Online]. Available:
  \url{http://dx.doi.org/10.14569/IJACSA.2018.090808}
\BIBentrySTDinterwordspacing

\bibitem{Li17}
\BIBentryALTinterwordspacing
X.~Li, R.~Li, M.~Wang, Y.~Liu, B.~Zhang, and J.~Zhou, ``Hyperspectral imaging
  and their applications in the nondestructive quality assessment of fruits and
  vegetables,'' in \emph{Hyperspectral Imaging in Agriculture, Food and
  Environment}, A.~I.~L. Maldonado, H.~R. Fuentes, and J.~A.~V. Contreras,
  Eds.\hskip 1em plus 0.5em minus 0.4em\relax Rijeka: IntechOpen, 2017, ch.~3.
  [Online]. Available: \url{https://doi.org/10.5772/intechopen.72250}
\BIBentrySTDinterwordspacing

\bibitem{article-bold}
B.~Boldrini, W.~Kessler, K.~Rebner, and R.~Kessler, ``Hyperspectral imaging: A
  review of best practice, performance and pitfalls for in-line and on-line
  applications,'' \emph{Journal of Near Infrared Spectroscopy}, vol.~20, pp.
  438--, 10 2012.

\bibitem{Lu2014MedicalHI}
G.~Lu and B.~Fei, ``Medical hyperspectral imaging: a review,'' \emph{Journal of
  Biomedical Optics}, vol.~19, 2014.

\bibitem{article-edel}
G.~Edelman, E.~Gaston, T.~van Leeuwen, P.~Cullen, and M.~Aalders,
  ``Hyperspectral imaging for non-contact analysis of forensic traces,''
  \emph{Forensic science international}, vol. 223, 10 2012.

\bibitem{8d8871b7b272427ea8b4d12e3e256ea4}
Z.~Luo, F.~Shafait, and A.~Mian, ``\BIBforeignlanguage{English}{Localized
  forgery detection in hyperspectral document images},'' in
  \emph{\BIBforeignlanguage{English}{2015 13th International Conference on
  Document Analysis and Recognition (ICDAR),}}, vol. n/a.\hskip 1em plus 0.5em
  minus 0.4em\relax United States: IEEE, Institute of Electrical and
  Electronics Engineers, 2015, pp. 496--500, localized forgery detection in
  hyperspectral document images ; Conference date: 01-01-2015.

\bibitem{10.1007/s11042-017-5564-2}
\BIBentryALTinterwordspacing
C.~Balas, G.~Epitropou, A.~Tsapras, and N.~Hadjinicolaou, ``Hyperspectral
  imaging and spectral classification for pigment identification and mapping in
  paintings by el greco and his workshop,'' \emph{Multimedia Tools Appl.},
  vol.~77, no.~8, p. 9737–9751, apr 2018. [Online]. Available:
  \url{https://doi.org/10.1007/s11042-017-5564-2}
\BIBentrySTDinterwordspacing

\bibitem{6628744}
Z.~Khan, F.~Shafait, and A.~Mian, ``Hyperspectral imaging for ink mismatch
  detection,'' in \emph{2013 12th International Conference on Document Analysis
  and Recognition}, 2013, pp. 877--881.

\bibitem{8270134}
A.~Abbas, K.~Khurshid, and F.~Shafait, ``Towards automated ink mismatch
  detection in hyperspectral document images,'' in \emph{2017 14th IAPR
  International Conference on Document Analysis and Recognition (ICDAR)},
  vol.~01, 2017, pp. 1229--1236.

\bibitem{article-kashif}
K.~Iqbal and K.~Khurshid, ``Automatic signature extraction from document images
  using hyperspectral unmixing,'' \emph{Proceedings of the Pakistan Academy of
  Sciences}, vol.~54, pp. 257--265, 09 2017.

\bibitem{8945886}
A.~U. Islam, M.~J. Khan, K.~Khurshid, and F.~Shafait, ``Hyperspectral image
  analysis for writer identification using deep learning,'' in \emph{2019
  Digital Image Computing: Techniques and Applications (DICTA)}, 2019, pp.
  1--7.

\bibitem{38d62691d8dc4a1a83da13335fde6bac}
S.~{Joo Kim}, F.~Deng, and M.~Brown, ``\BIBforeignlanguage{English}{Visual
  enhancement of old documents with hyperspectral imaging},''
  \emph{\BIBforeignlanguage{English}{Pattern Recognition}}, vol.~44, no.~7, pp.
  1461--1469, Jul. 2011, funding Information: We gratefully acknowledge the
  support and efforts from our collaborators Roberto Padoan from the Nationaal
  Archief of the Netherlands (NAN) and Marvin Klein from Art Innovation. This
  work was supported in part by the NUS Young Investigator Award,
  R-252-000-379-101.

\bibitem{Rahiche2020HistoricalDD}
A.~Rahiche, R.~Hedjam, S.~A. Al-Maadeed, and M.~Cheriet, ``Historical documents
  dating using multispectral imaging and ordinal classification,''
  \emph{Journal of Cultural Heritage}, vol.~45, pp. 71--80, 2020.

\bibitem{8978076}
M.~J. Khan, K.~Khurshid, and F.~Shafait, ``A spatio-spectral hybrid
  convolutional architecture for hyperspectral document authentication,'' in
  \emph{2019 International Conference on Document Analysis and Recognition
  (ICDAR)}, 2019, pp. 1097--1102.

\bibitem{article-khan}
M.~Khan, A.~Yousaf, A.~Abbas, and K.~Khurshid, ``Deep learning for automated
  forgery detection in hyperspectral document images,'' \emph{Journal of
  Electronic Imaging}, vol.~27, p. 053001, 09 2018.

\bibitem{Devassy2019InkCU}
B.~M. Devassy and S.~George, ``{Ink Classification Using Convolutional Neural
  Network},'' in \emph{X}, 2019.

\bibitem{10.1007/11669487_4}
D.~Fadoua, F.~Le~Bourgeois, and H.~Emptoz, ``Restoring ink bleed-through
  degraded document images using a recursive unsupervised classification
  technique,'' in \emph{Document Analysis Systems VII}, H.~Bunke and A.~L.
  Spitz, Eds.\hskip 1em plus 0.5em minus 0.4em\relax Berlin, Heidelberg:
  Springer Berlin Heidelberg, 2006, pp. 38--49.

\bibitem{article-qur}
R.~Qureshi, K.~Khurshid, and H.~Yan, ``Hyperspectral document image processing:
  Applications, challenges and future prospects,'' \emph{Pattern Recognition},
  vol.~90, pp. 12--22, 06 2019.

\bibitem{article-zohaib}
Z.~Khan, F.~Shafait, and A.~Mian, ``Automatic ink mismatch detection for
  forensic document analysis,'' \emph{Pattern Recognition}, vol.~48, 04 2015.

\bibitem{ISLAM2022107964}
\BIBentryALTinterwordspacing
A.~U. Islam, M.~J. Khan, M.~Asad, H.~A. Khan, and K.~Khurshid, ``ivision hhid:
  Handwritten hyperspectral images dataset for benchmarking hyperspectral
  imaging-based document forensic analysis,'' \emph{Data in Brief}, vol.~41, p.
  107964, 2022. [Online]. Available:
  \url{https://www.sciencedirect.com/science/article/pii/S2352340922001755}
\BIBentrySTDinterwordspacing

\end{thebibliography}

\end{document}